\documentclass[10pt,twocolumn,letterpaper, sigchi]{article}

\usepackage[pagenumbers]{cvpr} 

\usepackage{graphicx}
\usepackage{amsmath}
\usepackage{amssymb}
\usepackage{booktabs}
\usepackage{caption}
\usepackage{blindtext}
\usepackage{booktabs}
\usepackage[pagebackref,breaklinks,colorlinks]{hyperref}
\usepackage{floatrow}
\newfloatcommand{capbtabbox}{table}[][\FBwidth]

\usepackage[capitalize]{cleveref}
\crefname{section}{Sec.}{Secs.}
\Crefname{section}{Section}{Sections}
\Crefname{table}{Table}{Tables}
\crefname{table}{Tab.}{Tabs.}

\begin{document}

\title{FENeRF: Face Editing in Neural Radiance Fields}


\author{Jingxiang Sun$^{1}$\thanks{Work done during an internship at Tencent AI Lab.} \qquad Xuan Wang$^{2}$\thanks{Corresponding Author.} \qquad Yong Zhang$^{2}$ \qquad Xiaoyu Li$^{2}$ \\ Qi Zhang$^{2}$ \qquad Yebin Liu$^{3}$ \qquad Jue Wang$^{2}$ \vspace{5pt} \\
$^{1}$University of Illinois at Urbana-Champaign \qquad
$^{2}$Tencent AI Lab \qquad $^{3}$Tsinghua University \qquad\\
}

\maketitle

\begin{abstract}
\label{sec: abstract}

Previous portrait image generation methods roughly fall into two categories: 2D GANs and 3D-aware GANs. 2D GANs can generate high fidelity portraits but with low view consistency. 3D-aware GAN methods can maintain view consistency but their generated images are not locally editable. To overcome these limitations, we propose FENeRF, a 3D-aware generator that can produce view-consistent and locally-editable portrait images. Our method uses two decoupled latent codes to generate corresponding facial semantics and texture in a spatial-aligned 3D volume with shared geometry. Benefiting from such underlying 3D representation, FENeRF can jointly render the boundary-aligned image and semantic mask and use the semantic mask to edit the 3D volume via GAN inversion. We further show such 3D representation can be learned from widely available monocular image and semantic mask pairs. Moreover, we reveal that joint learning semantics and texture helps to generate finer geometry. Our experiments demonstrate that FENeRF outperforms state-of-the-art methods in various face editing tasks.

\end{abstract}

\section{Introduction}
\label{sec:intro}
Photo-realistic image synthesis via Generative Adversarial Networks is an important problem in computer vision and graphics. Specifically, synthesizing high-fidelity and editable portrait images has gained considerable attention in recent years. Two main classes of methods have been proposed: 2D GAN image generation and 3D-aware image synthesis techniques.

\begin{figure}[htbp]
    \centering
    
    \includegraphics[width=\textwidth]{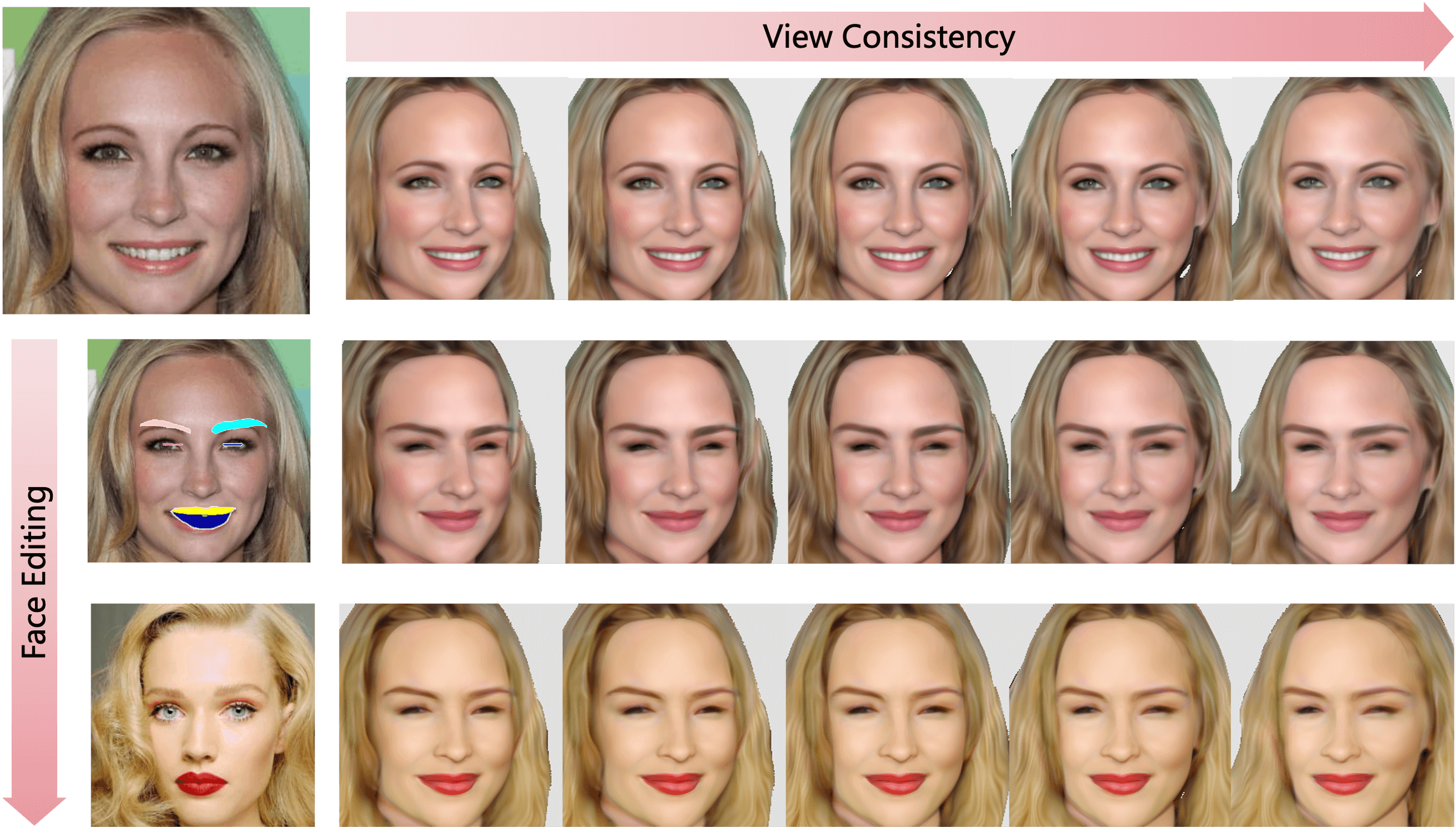}
    \caption{View-consistent portrait editing. Given the proposed FENeRF generator, we invert a given reference (top left) into shape and texture latent spaces to get the free-view portrait (first row). We can modify the rendered semantic masks and then leverage GAN inversion again to edit the free-view portraits (second row). As of last, we can replace the optimized texture code with the one from another image (bottom left) to perform the style transfer (bottom row) as well.}
    \label{fig: teaser}
\end{figure}

Despite their great success of synthesizing highly-realistic and even locally-editable images, 2D GAN methods all ignore the projection or rendering process of the underlying 3D scene, which is essential for view consistency. Consequently, they produce inevitable artifacts when changing the viewpoint of generated portraits. In order to overcome this issue, the Neural Radiance Fields (NeRF)~\cite{mildenhall2020nerf} have been explored to develop 3D-aware image synthesis techniques. Some of these methods~\cite{schwarz2020graf, chan2021pi} adopt vanilla NeRF generator to synthesize free-view portraits that are not editable, and the results could be blurry. Niemeyer et al.~\cite{niemeyer2021giraffe} employ volumetric rendering techniques to first produce view-consistency 2D feature maps, and then use an additional 2D decoder to obtain the final highly-realistic images. Nevertheless, such method suffers from additional view-dependent artifacts introduced by 2D convolutions and the mirror symmetry problem. To this end, CIPS-3D~\cite{zhou2021cips3d} replaces the 2D convolutions with implicit neural representation (INR) network. Unfortunately, all existing 3D-aware GANs do not support the interactive local editing on the generated free-view portraits.

In this paper, we propose a generator that can produce strictly view-consistent portraits, while supports interactive local editing. We adopt the noise-to-volume scheme. The generator takes as input the decoupled shape and texture latent code, and generates a 3D volume where the facial semantics and texture are spatially-aligned via the shared geometry. As a learnable 3D positional feature embedding is exploited while generating the texture volume, more details are preserved in the synthesized portraits. 

Directly learning this 3D volume representation is challenging due to the absent of suitable, large-scale 3D training data. A possible solution is to use multi-view images~\cite{chen2020sofgan}. Nonetheless, the inadequate training data harms the representation ability of the 3D semantic volume. To overcome this issue, we make the use of monocular images with paired semantic masks, which are vastly available.  Specifically, color and semantic discriminators are employed to supervise the training of the NeRF generator. The color discriminator focuses on image details hence improves the image fidelity. The semantic discriminator takes as input a pair of image and semantic map to enforce the alignment of corresponding content in the 3D volume. Thanks to the spatial-aligned 3D representation, we can use the semantic map to locally and flexibly edit the 3D volume via GAN inversion. In addition, an insight here is that learning the semantic and texture representations simultaneously helps to generate more accurate 3D geometry.

To illustrate the effectiveness of the proposed method, we perform the evaluation on two widely-used public datasets: CelebAMask-HQ and FFHQ. As shown in the experiments, the FENeRF generator outperforms state-of-the-art methods in several aspects. In addition, it supports various downstream tasks. To facilitate further research, we will release our code and models upon acceptance. To summarize, our main contributions are as following:

\begin{itemize}
    \item We present the first portrait image generator that is locally editable and strictly view-consistent, benefiting from the 3D representation in which the semantics, geometry and texture are spatially-aligned. 
    \item We train the generator with paired monocular images and semantic maps without the requirement of multi-view or 3D data. This ensures data diversity and enhances the representation ability of the generator.
    \item In experiments, we reveal that joint learning the semantic and texture volume can help to generate the finer 3D geometry.
\end{itemize}

\begin{figure*}[htbp]
    \centering
    \vspace{0.5cm}
    \includegraphics[width=\textwidth]{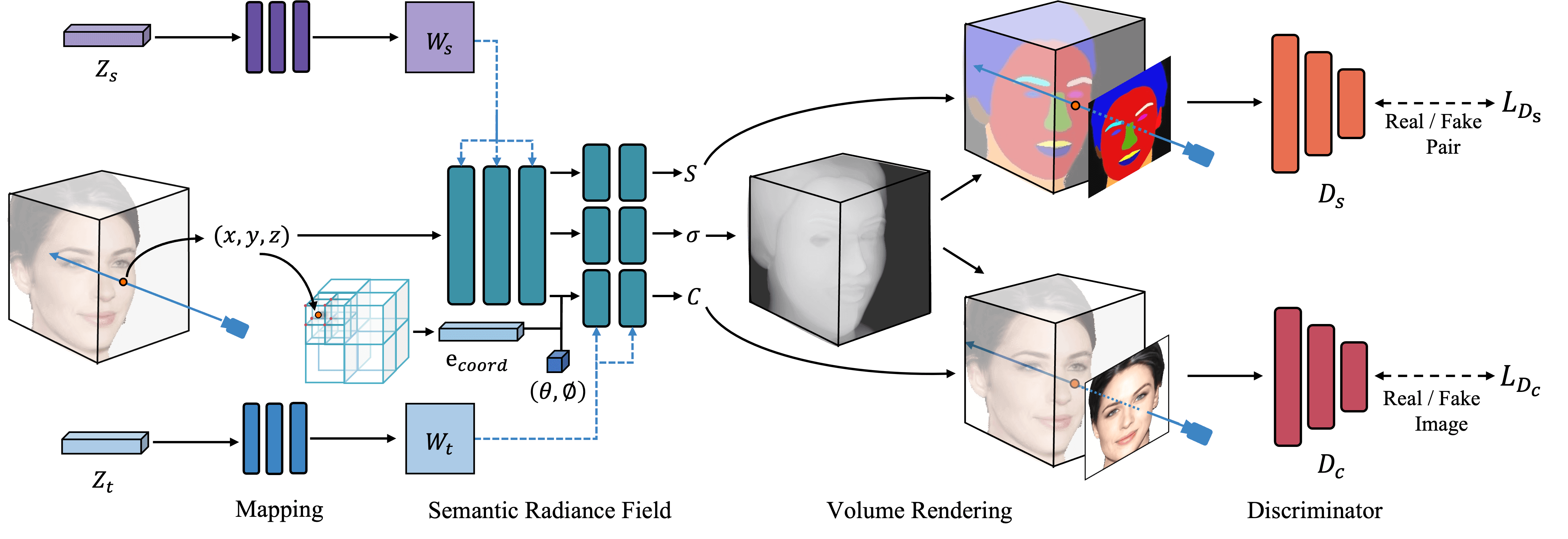} 
    \captionof{figure}{Overall pipeline of FENeRF. Our generator produces the spatially-aligned density, semantic and texture fields conditioned on disentangled latent codes $\mathbf{Z}_{s}$ and $\mathbf{Z}_{t}$. The positional feature embedding $e_{coord}$ is also injected to the network together with view direction for color prediction to preserve high-frequency details in generated image. By sharing the same density, aligned rgb image and semantic map are rendered. Finally two discriminators $D_s$ and $D_c$ are fed with semantic map/image pairs and real/fake image pairs, and trained with adversarial objectives $L_{D_s}$ and $L_{D_c}$, respectively.}
    \label{fig:pipeline}
\end{figure*}

\section{Related work}
\label{sec:Related Work}

\noindent \textbf{Neural Implicit Representations.}
Recently Neural implicit scene representation boosts various 3D perception tasks, such as 3D reconstruction and novel view synthesis, with its space continuity and memory efficiency. \cite{michalkiewicz2019implicit, mescheder2019occupancy, park2019deepsdf} represent either scene, objects as occupancy fields or signed distance functions, and 3D data is required for supervision. \cite{mildenhall2020nerf} models scene as neural radiance fields which are baked in weights of MLPs. With differentiable numerical integration of volume rendering, NeRF can be trained on only posed images. Various follow-ups extend NeRF to faster training and testing \cite{garbin2021fastnerf, reiser2021kilonerf, yu2021plenoctrees, hedman2021baking, cole2021differentiable}, pose-free \cite{lin2021barf, meng2021gnerf}, dynamic scenes \cite{yu2021pixelnerf, chen2021mvsnerf} and animating avatars \cite{peng2021animatable, liu2021neural, guo2021ad}. \cite{zhi2021place} extends NeRF with a semantic segmentation renderer and boosts performance of semantic interpretation. In this work, we build a generative semantic field aligned with neural radiance field. Instead of focusing on scene semantic understanding, we utilize the spatial alignment of facial texture and semantics to achieve semantic-guided attribute editing\vspace{1mm}. EditNeRF~\cite{liu2021editing} edits generic objects vis propagating 2D scribbles to the 3D space, nevertheless, only supports simple modifications on shape and color, while FENeRF tackles the complex facial deformation and detailed appearance texture, in the neural radiance fields, which are crucial for face editing task.

\noindent\textbf{Face Image Editing with 2D GANs.}
Generative Adversarial Networks (GANs) are widely used in photo-realistic face editing. Inspired by image-to-image translation, conditional GANs take as condition semantic masks \cite{fedus2018maskgan, isola2017image, park2019semantic, zhu2020sean} or hand-written sketches \cite{chen2020deepfacedrawing, li2020deepfacepencil, chen2021deepfaceediting} for the interactive editing of face images. SPADE \cite{park2019semantic} utilizes efficient spatially-adaptive normalization to synthesize photorealistic face images given input semantic layouts. SEAN \cite{zhu2020sean} further enables semantic region-based styling and more flexible facial editing. In order to provide explicit control of 3D interpretable semantic parameters (\eg pose, expression, illumination), several recent approaches decompose the image generation space into multiple specific attributes based on 3D guidance \cite{leimkuhler2021freestylegan, tewari2020stylerig, chen2020sofgan}. SofGAN proposes a semantic occupancy field to render view-consistent semantic maps which provide geometry constraints on image synthesis. However, SofGAN still lacks the interpretation of 3D geometry and considerable semantic labelled 3D scans are required for training semantic rendering. Instead, our FENeRF is trained end-to-end in an adversarial manner without any 3D data or multi-view images. Moreover, we show that our semantic rendering has better view consistency\vspace{1mm}.

\noindent\textbf{3D-Aware Image Synthesis.} Despite the tremendous breakthroughs in image generation by deep adversarial models \cite{karras2019style, karras2020analyzing, isola2017image, park2019semantic, zhu2020sean, collins2020editing, tewari2020stylerig}, those methods mainly manipulate shape and textures in 2D space without understanding of 3D nature of objects and scenes, resulting in limited pose control ability. To this end, 3D image synthesis methods lift image generation into 3D with explicit camera control. Early approaches \cite{zhu2018visual, gadelha20173d, nguyen2019hologan} utilize explicit voxel or volume representation thus are limited in resolution. Recently neural implicit scene representations are integrated to generative adversarial models and enable better memory efficiency and multi-view consistency\cite{chan2021pi, schwarz2020graf, niemeyer2021giraffe, eg3d, gram, gu2021stylenerf, orel2021stylesdf, zhou2021cips3d}. In particular, $\mathrm{\pi}$-GAN \cite{chan2021pi} presents siren-based neural radiance field conditioned on a global latent code which entangles geometry and texture. GRAF \cite{schwarz2020graf} and Giraffe \cite{niemeyer2021giraffe} enable disentangled control of texture and geometry, however, in a global level. The concurrent works~\cite{gu2021stylenerf, eg3d, gram, orel2021stylesdf, zhou2021cips3d} demonstrate impressive image generation quality but still don't support user-interacted local editing. By contrast, our FENeRF enables both global independent styling on texture and geometry as well as local facial attribute editing while preserving view consistency. 

\section{Method}


 
\subsection{Locally Editable NeRF Generator}
\label{sub: 3.1}
Our goal is to enable semantic-guided facial editing in 3D space. The main challenges are: 1) We need to decouple shape and texture during image generation. 2) Semantic map has to be strictly aligned with geometry and texture in 3D space. To this end, FENeRF exploits two separate latent codes. The shape latent code is to control the geometry and semantics. The texture code controls the appearance in the texture volume. Moreover, we exploits the three-head architecture in presented generator to individually encode the semantics and texture which are aligned with the underlying geometry depicted in the density volume. We formulate our generator as follows:

{\setlength\abovedisplayskip{1pt}
\begin{equation}
    \begin{split}
     G: 
     (\mathbf{x}, \mathbf{d}, \mathbf{z}_s, \mathbf{z}_t, \mathbf{e}_{coord}) \mapsto (\mathbf{\sigma}, \mathbf{c}, \mathbf{s}),
    \end{split}
\end{equation}}

As illustrated in Fig.~\ref{fig:pipeline}, the proposed generator is parameterized as Multi-Layer Perceptrons (MLPs), which takes as input the 3D point coordinates $\mathbf{x} = (x, y, z)$, viewing direction $\mathbf{d} = (\theta, \phi)$ and the learned positional feature embedding~$\mathbf{e}_{coord}$. Then generates the view-invariant density $\mathbf{\sigma} \in \mathbb{R}^+$ and semantic labels $\mathbf{s}_r \in \mathbb{R}^k$ conditioning on shape latent code~$\mathbf{z}_s$, as well as the view-dependent colour $\mathbf{c}_r \in \mathbb{R}^3$ conditioning on texture code~$\mathbf{z}_c$.

We also utilize a mapping network to map sampled codes into an intermediate latent space $\mathcal{W}$ and output frequencies $\mathbf{\gamma}$ and phase shifts $\mathbf{\beta}$, controlling the generator through feature-wise linear modulation as done in \cite{chan2021pi, perez2018film}:

\begin{align}
    \Phi(\mathbf{x})=\mathbf{W}_n(\phi_{n-1} \circ \phi_{n-2} \circ ... \circ \phi_{0})(\mathbf{x}) + \mathbf{b}_n \\
    \mathbf{x}_i \mapsto \phi_{i}(\mathbf{x}_i) = \mathrm{sin}(\mathbf{\gamma}_i \cdot (\mathbf{W}_i\mathbf{x}_i + \mathbf{b}_i) +  \mathbf{\beta}_i),
\end{align}

\noindent where $\phi_i: \mathbb{R}^{M_i} \mapsto \mathbb{R}^{N_i}$ is the $i^{th}$ layer of the network. The input $\mathbf{x}_i \in  \mathbb{R}^{M_i}$ is transformed by the weight matrix $\mathbf{W}_i \in \mathbb{R}^{N_i \times M_i}$ and the biases $\mathbf{b}_i \in \mathbb{R}^{M_i}$, and then modulated by the sine nonlinearity. 

Nevertheless, utilising the siren-based network only generates images lacking details. Therefore, we introduce a learnable 3D feature grid to compensate high-frequency image details. Specifically, to predict the color for a 3D point $\mathbf{x}$ with 2D view direction $\mathrm{d}$, we sample a local feature vector $\mathbf{e}_{coord}^{\mathrm{x}}$ from the feature grid by bi-cubic interpolation, then it is fed into the color branch as additional input. As shown in Fig.~\ref{fig:effect_fg}, it helps to preserve finer-grained image details.


Once the semantic, density and colour fields are generated, we can render those into semantic map and portrait image from arbitrary camera poses via volume rendering. For each 3D point, we first query its color $\mathbf{c}$, semantic labels $\mathbf{s}$ and volume density $\mathbf{\sigma}$. To obtain the pixel color $\mathbf{C}_r$ and semantic label probabilities $\mathbf{S}_r$, the values of all the samples in the ray are accumulated using the classical volume rendering process. The rendering equations are as follows:

\begin{equation}
    \label{eq: color rendering}
    \begin{aligned}
        \mathbf{C}(\mathbf{r})=\int_{t_n}^{t_f}T(t)\sigma(\mathbf{r}(t))\mathbf{c}(\mathbf{r}(t), \mathbf{d})dt, \\
    \end{aligned}
\end{equation}
\begin{equation}
    \label{eq: semantic rendering}
    \begin{aligned}
    \mathbf{S}(\mathbf{r})=\int_{t_n}^{t_f}T(t)\sigma(\mathbf{r}(t))\mathbf{s}(\mathbf{r}(t), \mathbf{d})dt, \\
    \end{aligned}
\end{equation}

\noindent where $T(t)=\mathrm{exp}(-\int_{t_n}^{t}\sigma(\mathbf{r}(s))ds).$ In practice, we approximate Eq.~\ref{eq: color rendering} and Eq.~\ref{eq: semantic rendering} in a discretized form following NeRF \cite{mildenhall2020nerf}. Note that the three branches of semantics, density and texture share the same intermediate features, and the output density is shared in the processes of colour and semantic rendering as well, to ensure that the generated semantics, density and texture are exactly aligned in 3D space.

\subsection{Discriminators}
\label{sub: 3.2}
In order to learn the unsupervised 3D representations, we design two discriminators $D_c$ and $D_s$ both of which are parameterized as CNN with leaky ReLU activation \cite{karras2017progressive}. $D_c$ discriminates the fidelity of generated portraits. Semantic masks, in addition to face images, are taken as input to $D_s$. This is to encourage the alignment of the face appearance and semantics. Moreover, we append two channels of $D_c$ to predict camera pose and then apply camera pose correction loss with sampled ones. 
\begin{figure}[t!]
    \centering
    \vspace{0.2cm}
    \includegraphics[width=\textwidth]{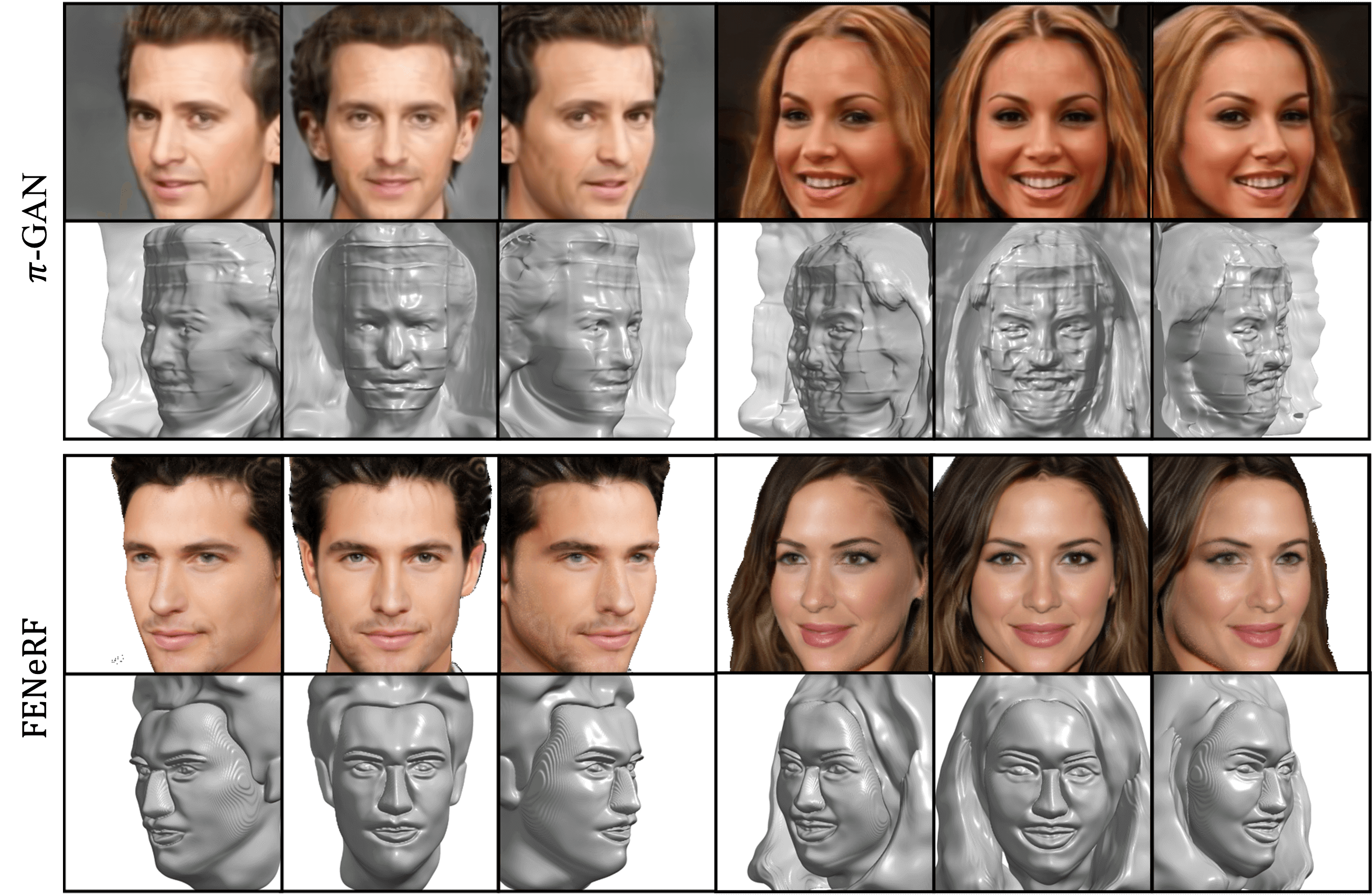}
    \caption{Comparison on geometry interpretation with $\mathrm{\pi}$-GAN. $\mathrm{\pi}$-GAN fails to learn accurate geometry (e.g. facial boundaries, hair, background), and suffers from serious artifacts. By contrast,
    benefiting from the semantic guidance, FENeRF generates accurate and smooth geometry without any specific regularization. Moreover, FENeRF enables a clear decouple of the generative 3D face from background.}
    \label{fig:geometry}
\end{figure}
\begin{figure}[htbp]
    \centering
    \includegraphics[width=\textwidth]{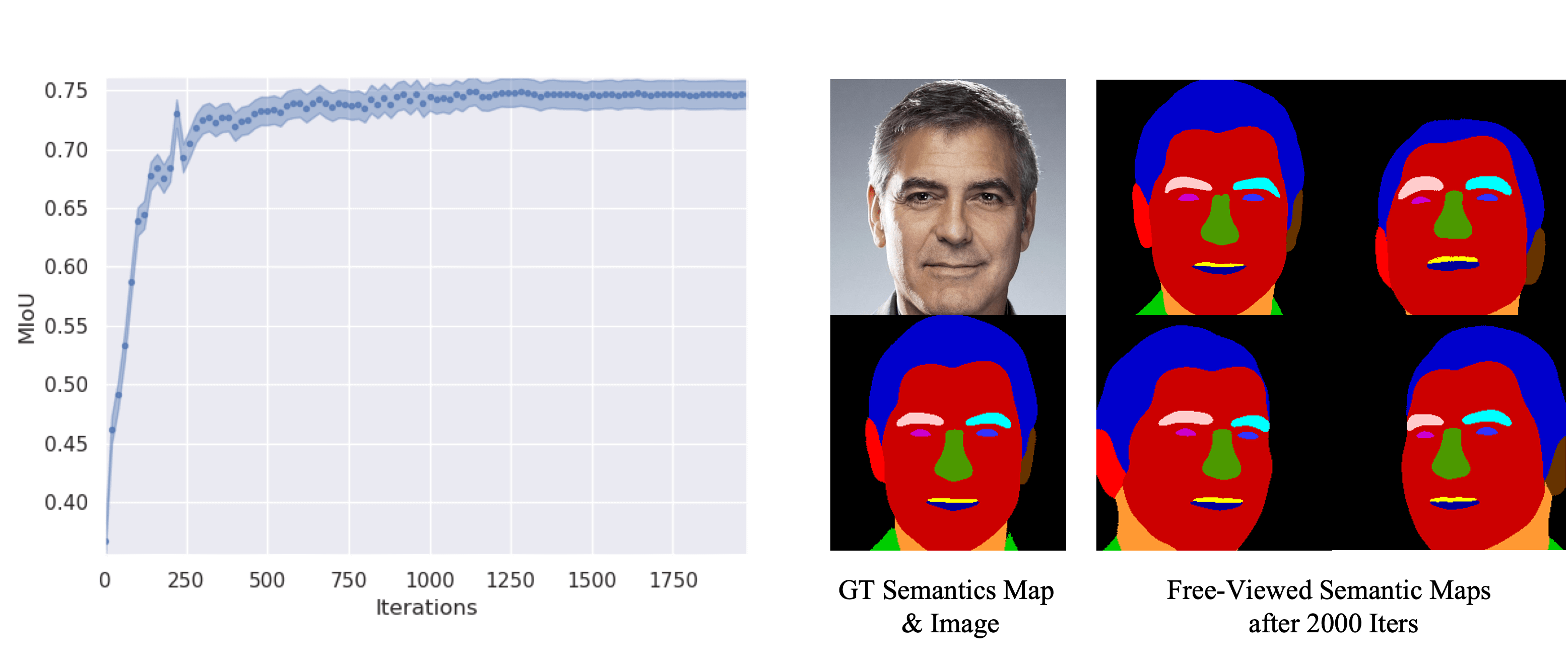}
    \caption{Average mIoU score during semantic inversion (left) and inversion results in free-view (right). We randomly select 1000 real face images from CelebAMask-HQ\cite{CelebAMask-HQ} and invert them into the shape latent space. The left chart illustrates the average mIoU score of the 1000 inverted semantic maps with respect to the iterations during semantic inversion. We visualize one reference portrait with its free-viewed inverted semantic maps on right.}
    \label{fig:miou}
\end{figure}

\subsection{Training}
\label{sub: 3.3}
During training, we randomly sample camera poses $\xi \sim p_{\xi}$ and latent codes $\mathbf{z}_s, \mathbf{z}_t \sim \mathcal{N}(0,I)$. We approximate the camera pose distribution as gaussian and set pose range as a prior according to \cite{schwarz2020graf, niemeyer2021giraffe, chan2021pi}. The camera positions are sampled on the surface of a object-centered sphere, and the camera is always directed towards the origin.

Our training loss is composed of three parts:
\begin{equation}
\vspace{0.2cm}
    \begin{aligned}
    \mathcal{L}_{D_c} = 
    &\mathbb{E}_{\mathbf{z}_s, \mathbf{z}_t \sim \mathcal{N}, \mathbf{\xi} \sim p_{\xi}}[f(D_{c}(\mathbf{x}_c))] + \\
    & \mathbb{E}_{\mathbf{I} \sim p_i}[f(-D_{c}(\mathbf{I})) + \\
    &\lambda_c\|\mathbf{\bigtriangledown}D_{c}(\mathbf{I})\|^2]
    \end{aligned}
    \label{eq:image discriminator loss}
\end{equation}
\begin{equation}
\vspace{0.2cm}
    \begin{aligned}
    \mathcal{L}_{D_s} = 
    &\mathbb{E}_{\mathbf{z}_s, \mathbf{z}_t \sim \mathcal{N}, \mathbf{\xi} \sim p_{\xi}}[f(D_{s}(\mathbf{x}_s, \mathbf{x}_c))] + \\ &\mathbb{E}_{\mathbf{I} \sim p_i, \mathbf{L} \sim p_l }[f(-D_{s}(\mathbf{L}, \mathbf{I})) + \\
    &\lambda_s\|\mathbf{\bigtriangledown}D_{s}(\mathbf{L}, \mathbf{I})\|^2] 
    \end{aligned}
    \label{eq:semantic discriminator loss}
\end{equation}
\begin{equation}
    \begin{split}
    \mathcal{L}_{G} = 
    &\mathbb{E}_{\mathbf{z}_s, \mathbf{z}_t \sim \mathcal{N}, \mathbf{\xi} \sim p_{\xi}}[f(D_{c}(\mathbf{x}_c))] + \\
    &\mathbb{E}_{\mathbf{z}_s, \mathbf{z}_t \sim \mathcal{N}, \mathbf{\xi} \sim p_{\xi}}[f(D_{s}(\mathbf{x}_s, \mathbf{x}_c))] + \\
    &\lambda_p\|\hat{\xi} - \xi\|
    \end{split}
    \label{eq:generator loss}
\end{equation}
\noindent where $f(t)=-\mathrm{log}(1+\mathrm{exp}(-t))$, $\mathbf{\lambda}_c, \mathbf{\lambda}_s, \mathbf{\lambda}_p =10$, and $p_i, p_l$ indicate the distributions of real images $\mathbf{I}$ and semantic maps $\mathbf{L}$ in datasets. The objectives of the image discriminator $D_c$,  semantic discriminator $D_s$ and generator $G$ to minimize $\mathcal{L}_{D_c}$, $\mathcal{L}_{D_s}$ and $\mathcal{L}_G$, respectively. $\mathcal{L}_{D_s}$ shown in Eq.~\ref{eq:semantic discriminator loss} is utilised to discriminate the paired image and semantic map and enforce their spatial alignment. While training generator $G$ with $\mathcal{L}_{G}$, we stop gradient back propagation from $D_s$ into color branch since the gradient would enforce texture to match semantics and lead to loss of fine image details. We adopt non-saturating GAN loss and $R_1$ gradient penalty\cite{mescheder2018training}. Moreover, we apply camera pose correction loss (the last term of Eq.~\ref{eq:generator loss}) to penalize the distance between camera pose $\hat{\xi}, \xi$ which is fed into generator and predicted by discriminator, respectively. This loss enforces all 3D faces lie in the same canonical pose and encourages a reliable 3D face geometry in avoid of pose drift.

In summary, our method builds a generative implicit representation which encodes facial geometry, texture and semantics jointly in a spatial aligned 3D volume. We further introduce a learnable feature grid for fine-grained image detail. An auxiliary discriminator further enforces this alignment by taking as input paired synthesised image and semantic map. Furthermore, we notice that semantic rendering significantly improves the quality of synthesised facial geometry as shown in Fig.~\ref{fig:geometry}.

\begin{figure}[t!]
    \centering
    \includegraphics[width=\textwidth]{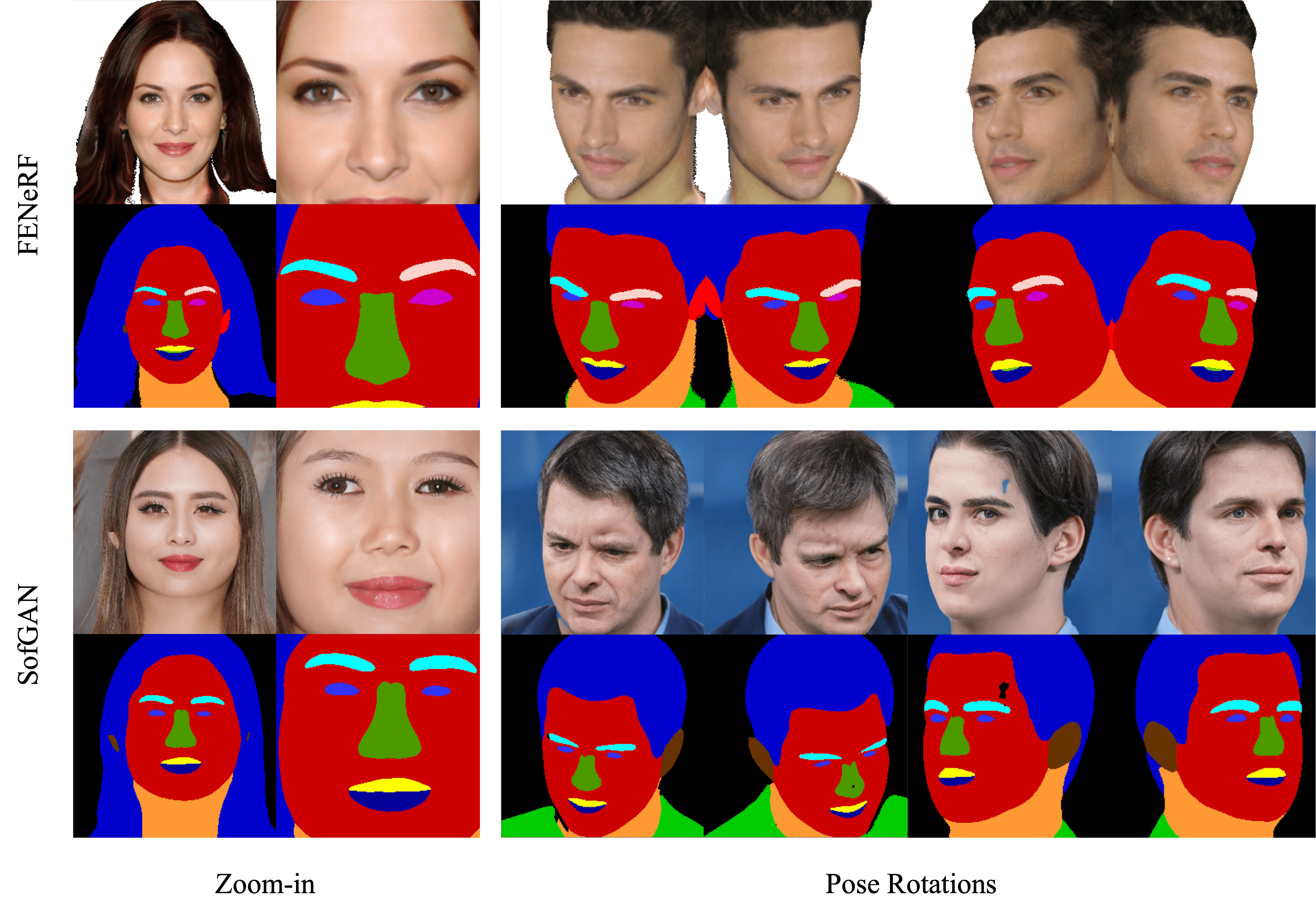}
    \caption{Comparison of semantic maps with SofGAN at extreme poses. SofGAN makes wrong semantic labels at extreme poses, leading to inconsistent artifacts in synthesized images. Besides, texture inconsistency (\eg inconsistent eye directions) appears when zooming in by SofGAN. Note that SofGAN is set to classify right and left attributes into the same class.}
    \label{fig:extreme_pose}
\end{figure}

\section{Experiments}
\label{sec: Experiments}

\label{sub: 4.1}
\noindent\textbf{Datasets.} 
We consider two datasets in our experiments for evaluation: CelebAMask-HQ~\cite{CelebAMask-HQ} and FFHQ~\cite{karras2019style}. CelebAMask-HQ contains 30,000 high-resolution face images from CelebA~\cite{liu2015deep} and each image has a segmentation mask of facial attributes. The masks have resolution of $512 \times 512$ and 19 classes including skin, eyebrows, ears, mouth, lip, etc. FFHQ contains 70,000 high-quality face images and We label the semantic classes by BiseNet \cite{yu2018bisenet}\vspace{2mm}. 

\noindent\textbf{Baselines.} We compare our model on image synthesis quality with three recent works for 3D-aware image synthesis : GRAF~\cite{schwarz2020graf}, pi-GAN\cite{chan2021pi} and Giraffe\cite{niemeyer2021giraffe}. We also compare the performance of semantic rendering with SofGAN~\cite{chen2020sofgan} which learns the semantic field with multi-view data and edits the portrait in 2D image plane. For the view consistency of inversion, we compare with InterfaceGAN\cite{shen2020interfacegan} and E4E\cite{tov2021designing}. We perform the evaluation of these approaches by leveraging their official implementations\vspace{2mm}.

\begin{figure}[htbp]
    \centering
    \vspace{-0.5cm}
    \includegraphics[width=\textwidth]{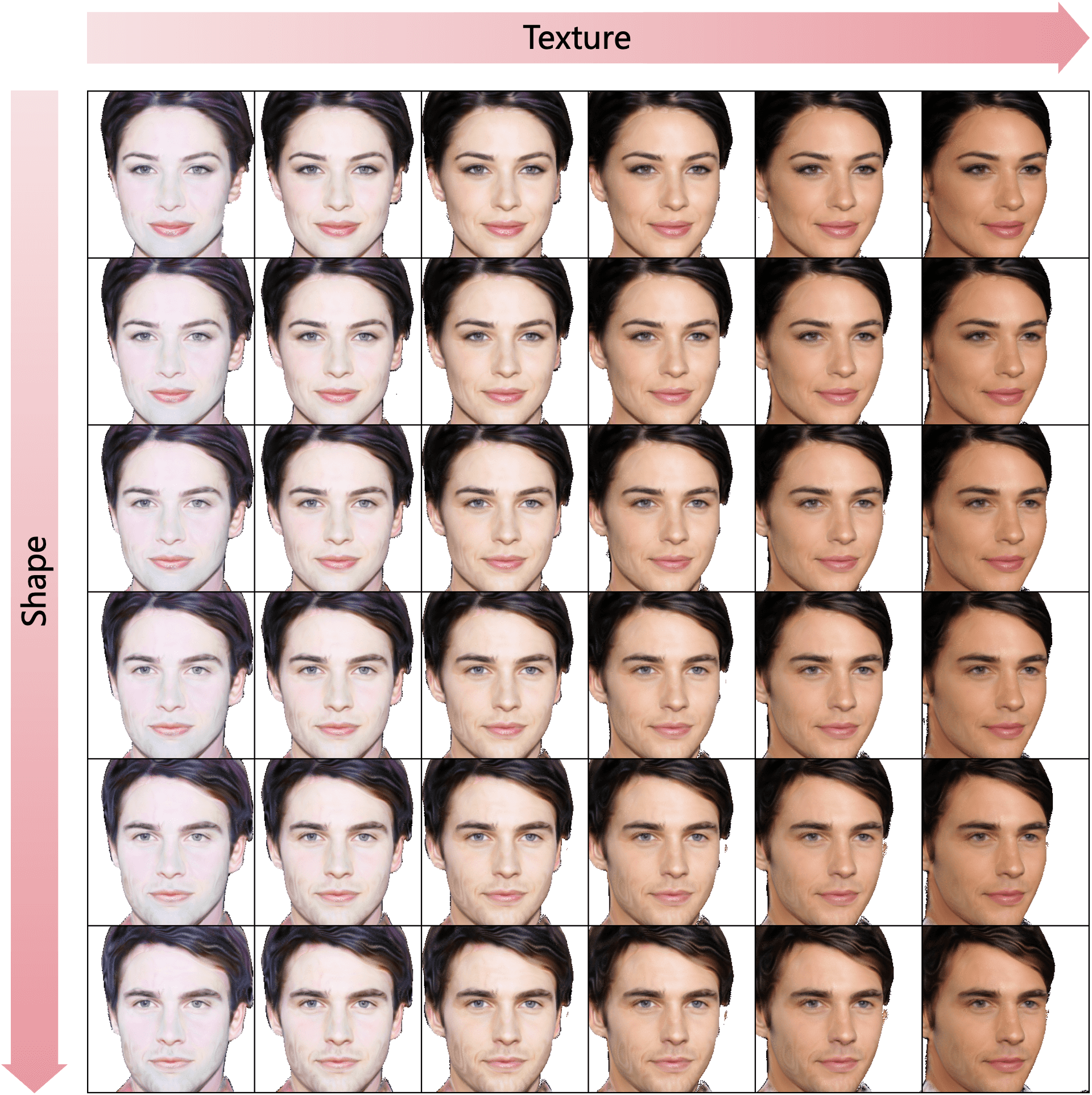}
    \caption{Visualization of disentangled morphing. We interpolate the texture and shape codes of images at the left bottom and right top. In each row, the texture code is interpolated while keeping the shape code constant. Similarly, the shape code varies across columns. We can see that the interpolated results are disentangled clearly in these two dimensions.}
    \label{fig:morphing}
\end{figure}

\begin{figure*}[!h]
    \centering
    \includegraphics[width=\textwidth]{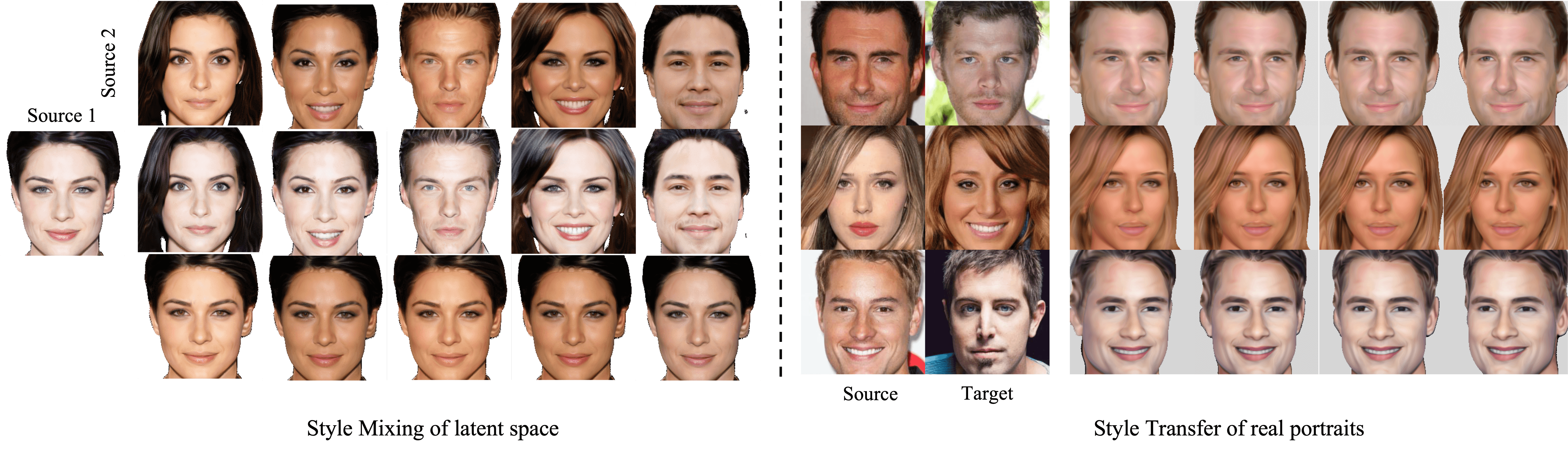}
    \caption{Global style editing. Style mixing (on left part): Given the Source 1 image and a sequence of source 2 images (on top row), the mixed face image in second row retains texture from source 1 and shape from source 2. Inversely, we show the mixed results with shape of source 1 and texture of source 2 sequence on bottom. Style transfer (on the right part): Given source and target images, we can also transfer the texture style of source image into the target one, and the free-view portraits are generated while facial geometry is preserved.}
    \label{fig: style transfer}
\end{figure*}

\noindent\textbf{Implementation details.} 
Our semantic radiance field is parameterized as MLPs with FiLM-conditioned SIREN layers~\cite{chan2021pi}. For discriminator, CoordConv layers~\cite{liu2018intriguing}, residual connections~\cite{he2016deep} are utilized. During training, we initialize the learning rate to $6 \times 10^{-5}$ for generator $G$, $2 \times 10^{-4}$ for image discriminator $D_c$ and $1 \times 10^{-4}$ for segmentation discriminator $D_s$. We use the Adam optimizer with $\beta_1 = 0, \beta_2 = 0.9$. We start training at $32 \times 32$  with a batch size of 40. Then the resolution is increased to $128 \times 128$ with a batch size of 24. Please refer to the supplemental material for more details.

\subsection{Comparisons}
\noindent\textbf{Quality evaluation of synthesized image.} 
We conduct quantitative comparisons of synthesized images with the state-of-the-art 3D-aware GAN methods and the results are shown in Table~\ref{tab:comparison}. Frechet Inception Distance (FID) \cite{heusel2017gans} and Kernel Inception Distance \cite{binkowski2018demystifying} are used to evaluate the image quality. For fair comparison, we retrain all models on these two datasets with full images (30k for CelebA-HQ and 70k for FFHQ) at $128 \times 128$ resolution. FID score is calculated 2048 randomly sampled images. We reach the state-of-the-art performance on both datasets with FID and KID. This improvement is attributed to: 1) the joint learning with semantic field provides reliable semantic enforcement and facilitate training converge; 2) our learnable feature grid brings high frequency details to synthesized images\vspace{2mm}.
\begin{table}
\resizebox{\linewidth}{!}{
\begin{tabular}{c|cc|cc}
\hline
\multicolumn{1}{l|}{} & \multicolumn{2}{c|}{\textbf{FID}$\downarrow$} & \multicolumn{2}{c}{\textbf{KID} ($\times 10^3$)$\downarrow$} \\ \cline{2-5}
& CelebA-HQ & FFHQ & CelebA-HQ & FFHQ     \\ \hline
GRAF    & 34.7 & 66.5 & 15.6 & 49.3  \\
pi-GAN  & 14.7 & 40.3 & 3.9 & 23.5  \\
Giraffe & 16.2 & 31.9 & 9.1 & 32.7  \\
FENeRF  & \textbf{12.1} & \textbf{28.2} & \textbf{1.6} & \textbf{17.3}  \\ \hline
\end{tabular}}
\caption{Quantitative comparison of our approach with other 3D-aware GAN methods. Our method outperforms other methods in terms of FID and KID on both CelebA-HQ and FFHQ datasets.}
\label{tab:comparison}
\end{table}

\noindent\textbf{Rendering performance of Semantic field.}
In our framework, we construct a neural semantic radiance field for a generative 3D face and render both semantic maps and images from arbitrary view points. To evaluate the performance of semantic rendering, we compare with SofGAN~\cite{chen2020sofgan} which trains a generative semantic occupancy field supervised by labeled multi-view semantic maps. As Fig.~\ref{fig:extreme_pose} shows, SofGAN is prone to wrong semantic classes at extreme poses and causes artifacts in synthesized face. We attribute the semantic inconsistency of SofGAN to that its semantic rendering relies on the surface construction which is, however, highly ambiguous from pure semantic maps even with multi-view observations. By contrast, FENeRF learns accurate geometry (Fig.~\ref{fig:geometry}) benefiting from the joint semantic and image rendering thus keeps view consistency at such challenging poses. Besides, for SofGAN, though the semantic map keeps consistent during zooming in, its  synthesised images are not view consistent (\eg eyes look into different directions). By contrast, FENeRF guarantees strict pixel-level view consistency of synthesised images. 

To further explore the inversion ability of semantic rendering, we randomly collect 1000 real portrait images and reproject them into the geometry latent space to recover a semantic field. The left chart in Fig.~\ref{fig:miou} illustrates that the average mIoU score of these 1000 portraits reaches 0.5 by 100 iterations and finally converges to over 0.7 within 200 iterations. This indicates the geometry latent space of FENeRF covers various portrait shapes. From the visualized example we can see that the facial semantics are reconstructed accurately after 2000 iterations with texture-aligned region boundaries and view consistency.

\subsection{Applications}
\label{sub: 4.3} 
\noindent \textbf{Disentangled morphing and style mixing.} Our method enables disentangled control on geometry and texture through independent latent sampling. Fig.~\ref{fig:morphing} demonstrates the disentangled morphing along two independent directions. Specifically , we sample two sets of texture and shape codes $(\mathrm{Z}^1_t, \mathrm{Z}^1_s)$, $(\mathrm{Z}^2_t, \mathrm{Z}^2_s)$ which synthesize images $\mathrm{I}^1, \mathrm{I}^2$ at left bottom and right top corners of Fig.~\ref{fig:morphing}. Then we perform linear interpolation in these two latent directions and group them for image synthesis. Our approach also supports style mixing demonstrated in Fig.~\ref{fig: style transfer} (on left), proving the effectiveness of our disentangled facial representation\vspace{2mm}.

\begin{figure*}[!h]
    \centering
    \includegraphics[width=\textwidth]{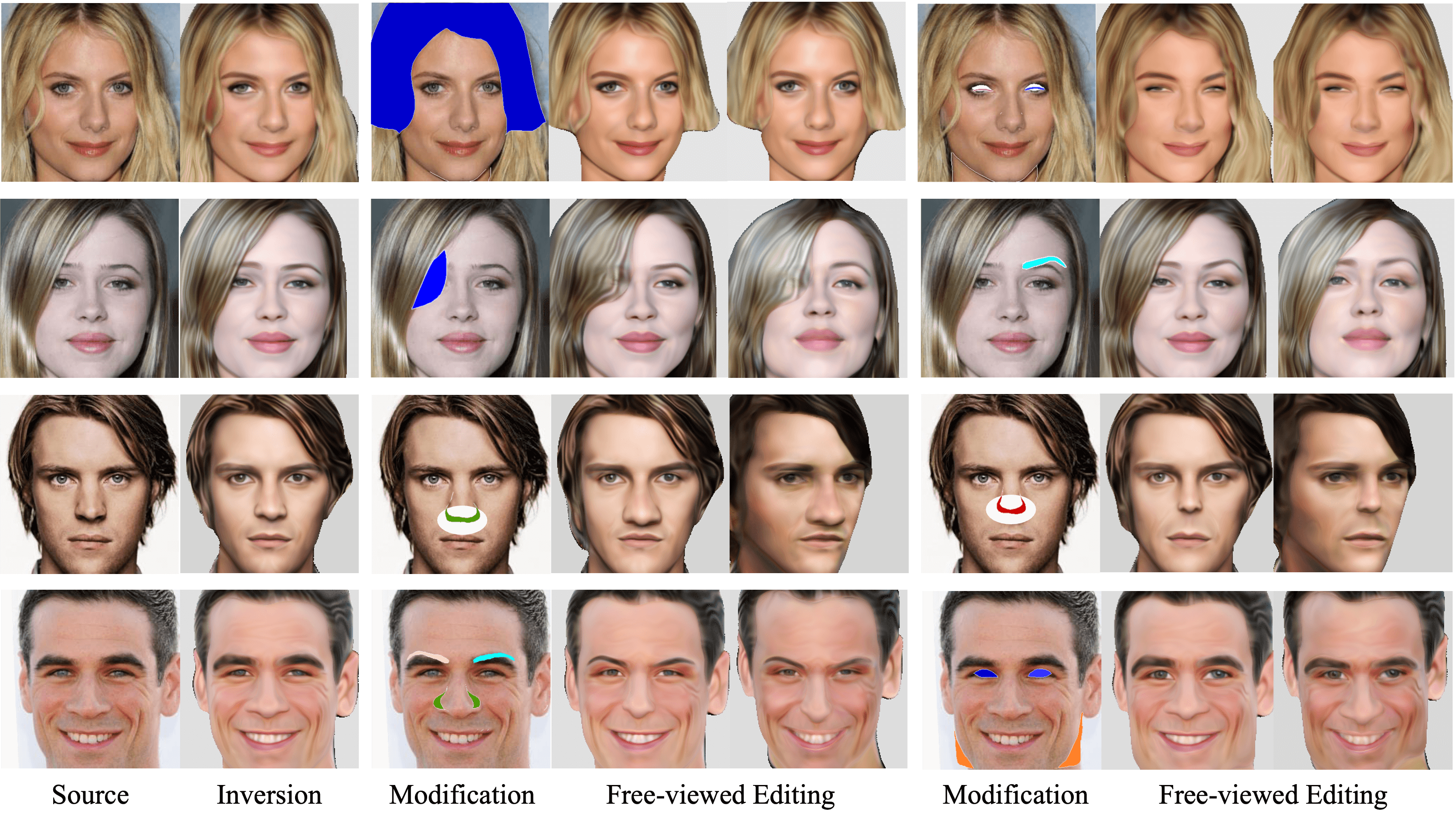}
    \caption{\textbf{Interactive image manipulation.} Our method enables interactive image editing with semantic guidance. Given the source image, we invert it into texture and shape codes and obtain the inversion results. We manipulate facial attributes on the semantic map (e.g. haircut, eyes, nose and face shape, etc.) and leverage the GAN inversion again. We obtain the corresponding modified free-view portraits.}
    \label{fig: local_editing}
\end{figure*}

\begin{figure}[!h]
    \centering
    \captionsetup{type=figure}
    \includegraphics[width=\textwidth]{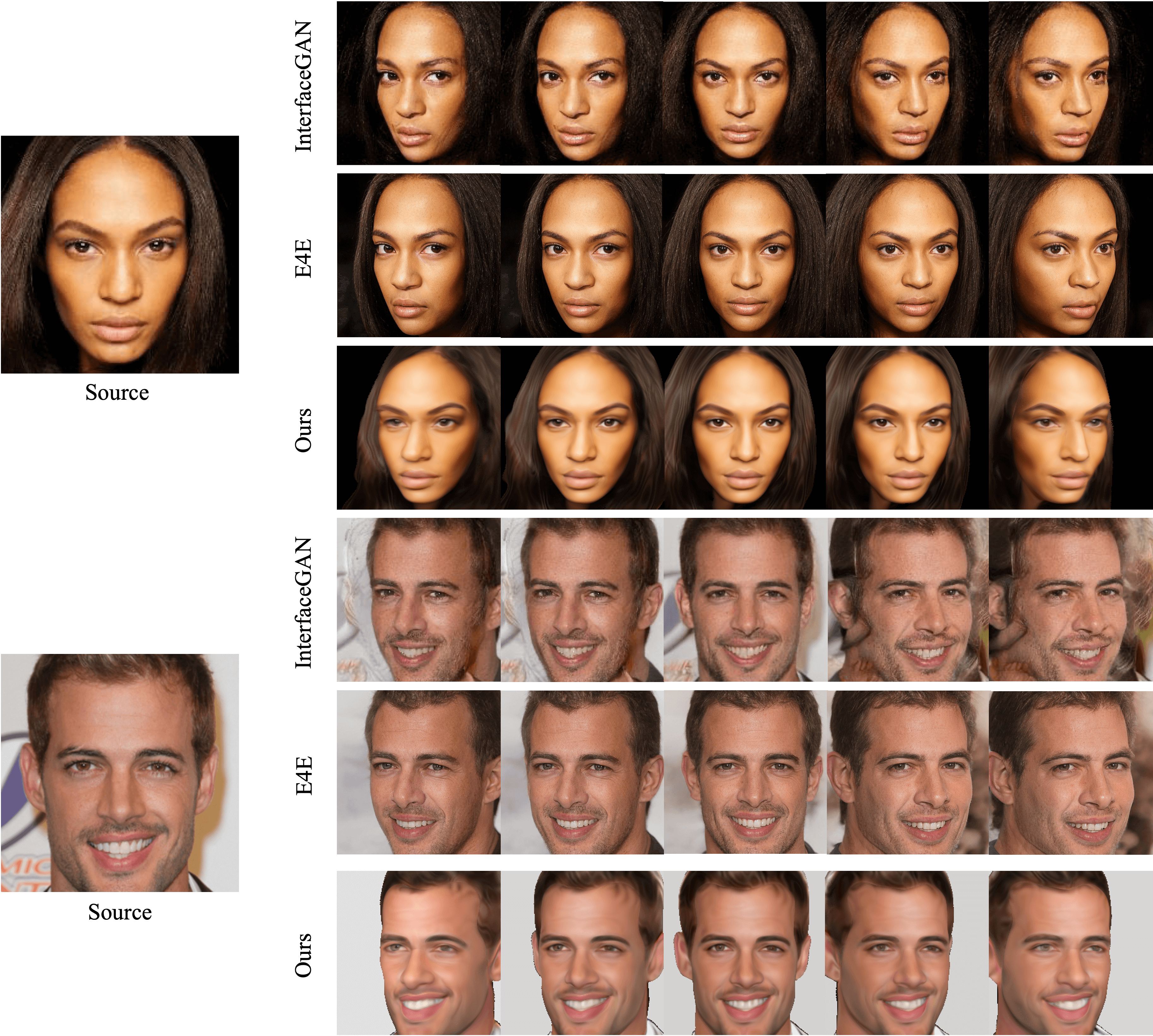}
    \caption{View consistency of inversion. We compare FENeRF against SOTA methods, InterFaceGAN\cite{shen2020interfacegan} and E4E\cite{tov2021designing}. Obviously, when generated portraits rotate,  InterFaceGAN suffers from the serious artifacts and E4E fails to preserve the face identity.}
    \label{fig: view consistency}
\end{figure}

\noindent\textbf{3D inversion and style transfer.} Inversion is a popular application of 2D GANs and we further lift it to 3D space and render in arbitrary poses. Though some recent conditional GANs \cite{leimkuhler2021freestylegan, shen2020interfacegan, wang2021high, tov2021designing} also support pose rotation after inversion, however, suffer from inconsistent identity and texture during rotation. Fig.~\ref{fig: view consistency} compares FENeRF with InterfaceGAN and E4E on realistic inversion with pose rotations. InterfaceGAN reconstructs realistic synthesized image for input view but generates obvious artifacts (e.g. sticking texture), while E4E generates inconsistent shape during camera rotations. Compared with 2D GANs, FENeRF outperforms these methods on view consistency of identity and texture with a slight drop of texture details. This is because we reproject the 2D image into a latent space which bounds to a 3D generative volume rather than a 2D space. Therefore, the latent code controls facial properties independent of camera pose. We also support face style transfer which is shown in Fig.~\ref{fig: style transfer}. Style transfer is more challenging than interpolation or mixing since real portraits' texture and shape could be far from the distribution of the latent spaces, leading to unrealistic artifacts. As shown in Fig.~\ref{fig: style transfer},  FENeRF successes to transfer the target texture to the source identity. Moreover, the results of style transfer holds consistency at poses far from the input views\vspace{2mm}.



\noindent \textbf{3D local editing.} Both style mixing and transfer manipulate face in a global manner and prove our powerful disentangled control of global texture and geometry. We further find that the semantic field would encourage the latent spaces of geometry and texture to be regional disentangled based on the spatial-aligned 3D volume. To prove that, we take facial attributes manipulation by editing semantic maps interactively. Fig.~\ref{fig: local_editing} demonstrates the results of facial attributes editing. We first inverse a real face image into a reconstructed image and the accompanied semantic map. Then we edit this semantic map and reproject it into geometry latent space by optimizing the shape code. Note FENeRF is capable of manipulating facial attributes even with significant deformation (\eg the enlarged and shorten noses in Fig.~\ref{fig: local_editing}) while keeping other regions consistent in shape and texture. Moreover, the edited faces by FENeRF still hold strict view consistency\vspace{2mm}.

\subsection{Ablation Studies}
\label{sub: 4.4}

\begin{figure}[t!]
    \centering
    \includegraphics[width=\textwidth]{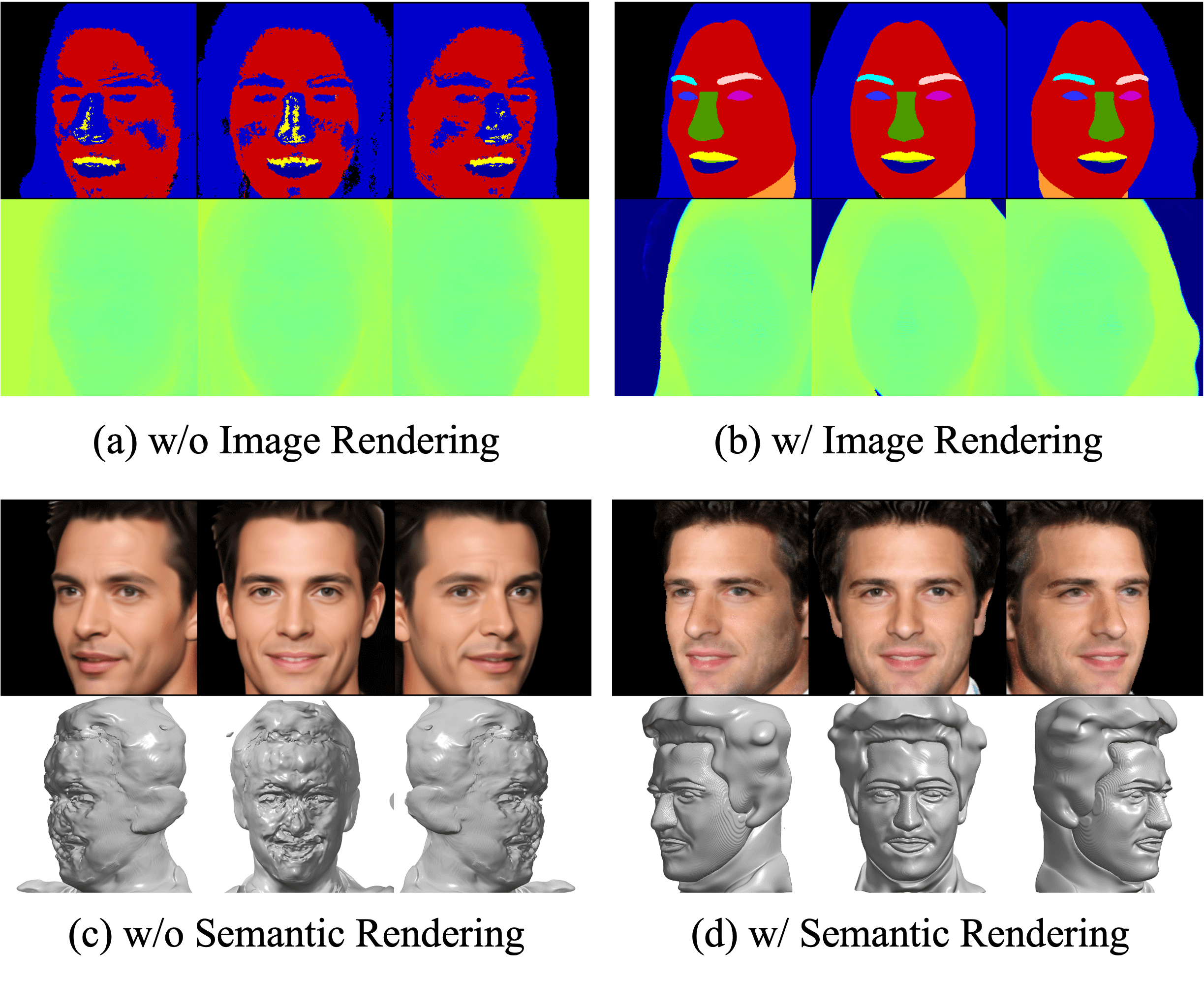}
    \caption{Effect of the joint semantic and image rendering. We show the effect of image rendering with rendered semantic maps (top row) and depth maps (bottom row) in the sub-page (a) and (b), and the effect of semantic rendering with rendered images(top row) and extracted meshes(bottom row)  in sub-pages (c) and (d). It is illustrated that, on one hand, learning the semantic and geometry fields with semantic map supervision only is impossible. On the other, our joint rendering enhances the quality of semantic, geometry and texture at the same time.}
    \label{fig:effect_semantic_rendering}
\end{figure}

\noindent \textbf{Benefits of our joint framework.} 
Recall that FENeRF renders face images accompanied by semantic maps. We conduct experiments to explore if this joint framework benefits two kinds of rendering. We first train a FeNeRF without image rendering. Therefore, given a 3D point, we only query its density and semantic labels with rendering a single semantic map. In Fig.~\ref{fig:effect_semantic_rendering}, FENeRF without image rendering in (a) fails to specify facial regions or converge to reliable geometry. By contrast, training with image rendering in (b) solve these problems. We further explore effect of semantic rendering on geometry quality. (c) and (d) in Fig.~\ref{fig:effect_semantic_rendering} demonstrates that semantic rendering encourages the generative surface to smooth and accurate. This result is consistent with Fig.~\ref{fig:geometry}\vspace{2mm}.

\begin{figure}[htbp]
    \centering
    \includegraphics[width=\textwidth]{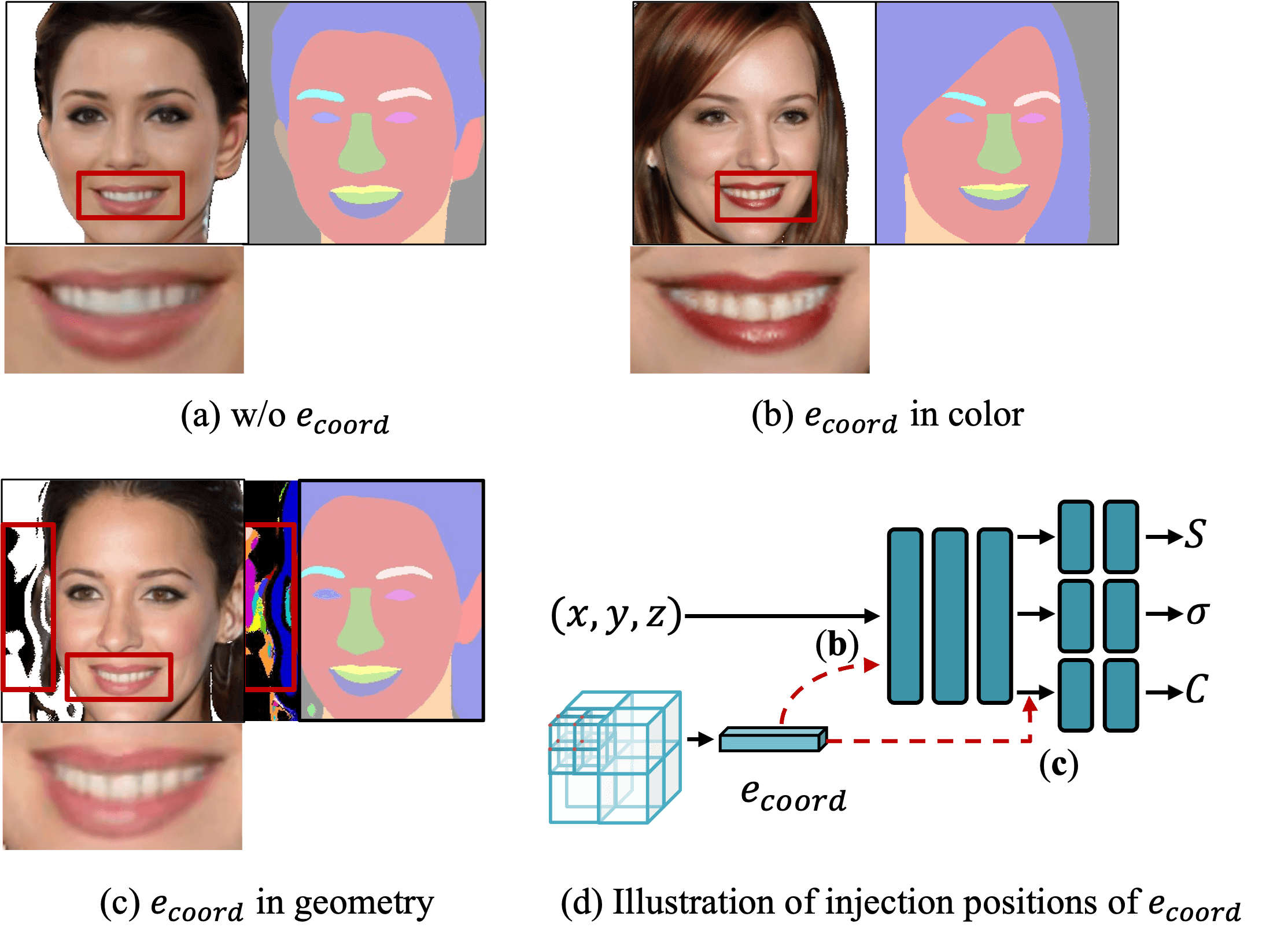}
    \caption{Ablation study on $e_{coord}$. Sub pages (a) to (c) are the generated image details with three variants of $e_{coord}$. We zoom in the synthesized mouths for more clear observation. (d) illustrates the injecting positions of $e_{coord}$ in (b) and (c).}
    \label{fig:effect_fg}
\end{figure}

\noindent \textbf{Effects of learnable coordinate embeddings.}
To produce fine-grained image details, we introduce a learnable feature grid for the local sampling of coordinate embedding ($e_{coord}$). We further conduct experiments to explore the effect of $e_{coord}$ and its injecting positions. As shown in Fig.~\ref{fig:effect_fg}, FENeRF w/o $e_{coord}$ in (a) generates blurry teeth. Injecting $e_{coord}$ as in (b) generates sharper details but suffers from artifacts on synthesised images and semantic map since $e_{coord}$ brings high frequency signal into the volume density when injecting at the beginning of MLP with coordinates. (c) injects $e_{coord}$ into the color branch and enables both high-frequency image and smooth semantics\vspace{2mm}.

\section{Limitations}
One known limitation is our generator cannot produce HD portrait images due to the computationally-expensive ray casting and volume integration. 
Besides, GAN inversion is an effective method to perform the local editing of 3D volume. The iterative optimization of inversion, however, is inefficient. As a result, the real-time free-view portrait editing is still an open problem.


\section{Conclusion} 
In this paper we present the first locally editable 3D-aware face generator FENeRF based on implicit scene representation. For using semantic map as editing interface, we introduce a semantic radiance field which aligns facial semantics and texture implicitly in 3D space through the shared geometry. We show that FENeRF realizes fancy applications including style mixing, style transfer, facial attribute editing, and we further push them to a 3D free-view manner with explicit camera control. We hope our work would introduce a promising research direction of editable 3D-aware generative network. For future works, we plan to increase the resolution of synthesised free-view portraits and study the specific 3D-aware GAN inversion as well. 

\section{Potential Social Impact} 
Given a single real portrait image, FENeRF enables the generation of his/her photo-realistic avatar through GAN inversion. Moreover, we can drive this avatar by changing semantic maps and camera poses to make fake videos. Therefore there are certain risks of fooling face recognition systems, such as vivo detection, with our synthesised fake videos thus it should be careful to deploy the technology.

{\small
\bibliographystyle{ieee_fullname}
\bibliography{egbib}

\begin{thebibliography}{10}\itemsep=-1pt

\bibitem{binkowski2018demystifying}
Miko{\l}aj Bi{\'n}kowski, Danica~J Sutherland, Michael Arbel, and Arthur
  Gretton.
\newblock Demystifying mmd gans.
\newblock {\em arXiv preprint arXiv:1801.01401}, 2018.

\bibitem{eg3d}
Eric~R. Chan, Connor~Z. Lin, Matthew~A. Chan, Koki Nagano, Boxiao Pan,
  Shalini~De Mello, Orazio Gallo, Leonidas Guibas, Jonathan Tremblay, Sameh
  Khamis, Tero Karras, and Gordon Wetzstein.
\newblock Efficient geometry-aware {3D} generative adversarial networks.
\newblock In {\em arXiv}, 2021.

\bibitem{chan2021pi}
Eric~R Chan, Marco Monteiro, Petr Kellnhofer, Jiajun Wu, and Gordon Wetzstein.
\newblock pi-gan: Periodic implicit generative adversarial networks for
  3d-aware image synthesis.
\newblock In {\em Proceedings of the IEEE/CVF Conference on Computer Vision and
  Pattern Recognition}, pages 5799--5809, 2021.

\bibitem{chen2020sofgan}
Anpei Chen, Ruiyang Liu, Ling Xie, Zhang Chen, Hao Su, and Jingyi Yu.
\newblock Sofgan: A portrait image generator with dynamic styling.
\newblock {\em ACM Transactions on Graphics (TOG)}, 41(1):1--26, 2022.

\bibitem{chen2021mvsnerf}
Anpei Chen, Zexiang Xu, Fuqiang Zhao, Xiaoshuai Zhang, Fanbo Xiang, Jingyi Yu,
  and Hao Su.
\newblock Mvsnerf: Fast generalizable radiance field reconstruction from
  multi-view stereo.
\newblock In {\em Proceedings of the IEEE/CVF International Conference on
  Computer Vision}, pages 14124--14133, 2021.

\bibitem{chen2021deepfaceediting}
Shu-Yu Chen, Feng-Lin Liu, Yu-Kun Lai, Paul~L. Rosin, Chunpeng Li, Hongbo Fu,
  and Lin Gao.
\newblock Deepfaceediting: Deep generation of face images from sketches.
\newblock {\em ACM Transactions on Graphics (TOG)}, 40(4):90:1--90:15, 2021.

\bibitem{chen2020deepfacedrawing}
Shu-Yu Chen, Wanchao Su, Lin Gao, Shihong Xia, and Hongbo Fu.
\newblock Deepfacedrawing: Deep generation of face images from sketches.
\newblock {\em ACM Transactions on Graphics (TOG)}, 39(4):72--1, 2020.

\bibitem{cole2021differentiable}
Forrester Cole, Kyle Genova, Avneesh Sud, Daniel Vlasic, and Zhoutong Zhang.
\newblock Differentiable surface rendering via non-differentiable sampling.
\newblock In {\em Proceedings of the IEEE/CVF International Conference on
  Computer Vision}, pages 6088--6097, 2021.

\bibitem{collins2020editing}
Edo Collins, Raja Bala, Bob Price, and Sabine Susstrunk.
\newblock Editing in style: Uncovering the local semantics of gans.
\newblock In {\em Proceedings of the IEEE/CVF Conference on Computer Vision and
  Pattern Recognition}, pages 5771--5780, 2020.

\bibitem{gram}
Yu Deng, Jiaolong Yang, Jianfeng Xiang, and Xin Tong.
\newblock Gram: Generative radiance manifolds for 3d-aware image generation.
\newblock In {\em arXiv}, 2021.

\bibitem{fedus2018maskgan}
William Fedus, Ian Goodfellow, and Andrew~M Dai.
\newblock Maskgan: better text generation via filling in the\_.
\newblock {\em arXiv preprint arXiv:1801.07736}, 2018.

\bibitem{gadelha20173d}
Matheus Gadelha, Subhransu Maji, and Rui Wang.
\newblock 3d shape induction from 2d views of multiple objects.
\newblock In {\em 2017 International Conference on 3D Vision (3DV)}, pages
  402--411. IEEE, 2017.

\bibitem{garbin2021fastnerf}
Stephan~J Garbin, Marek Kowalski, Matthew Johnson, Jamie Shotton, and Julien
  Valentin.
\newblock Fastnerf: High-fidelity neural rendering at 200fps.
\newblock In {\em Proceedings of the IEEE/CVF International Conference on
  Computer Vision}, pages 14346--14355, 2021.

\bibitem{gu2021stylenerf}
Jiatao Gu, Lingjie Liu, Peng Wang, and Christian Theobalt.
\newblock Stylenerf: A style-based 3d-aware generator for high-resolution image
  synthesis, 2021.

\bibitem{guo2021ad}
Yudong Guo, Keyu Chen, Sen Liang, Yong-Jin Liu, Hujun Bao, and Juyong Zhang.
\newblock Ad-nerf: Audio driven neural radiance fields for talking head
  synthesis.
\newblock In {\em Proceedings of the IEEE/CVF International Conference on
  Computer Vision}, pages 5784--5794, 2021.

\bibitem{he2016deep}
Kaiming He, Xiangyu Zhang, Shaoqing Ren, and Jian Sun.
\newblock Deep residual learning for image recognition.
\newblock In {\em Proceedings of the IEEE conference on computer vision and
  pattern recognition}, pages 770--778, 2016.

\bibitem{hedman2021baking}
Peter Hedman, Pratul~P Srinivasan, Ben Mildenhall, Jonathan~T Barron, and Paul
  Debevec.
\newblock Baking neural radiance fields for real-time view synthesis.
\newblock In {\em Proceedings of the IEEE/CVF International Conference on
  Computer Vision}, pages 5875--5884, 2021.

\bibitem{heusel2017gans}
Martin Heusel, Hubert Ramsauer, Thomas Unterthiner, Bernhard Nessler, and Sepp
  Hochreiter.
\newblock Gans trained by a two time-scale update rule converge to a local nash
  equilibrium.
\newblock {\em Advances in neural information processing systems}, 30, 2017.

\bibitem{isola2017image}
Phillip Isola, Jun-Yan Zhu, Tinghui Zhou, and Alexei~A Efros.
\newblock Image-to-image translation with conditional adversarial networks.
\newblock In {\em Proceedings of the IEEE conference on computer vision and
  pattern recognition}, pages 1125--1134, 2017.

\bibitem{karras2017progressive}
Tero Karras, Timo Aila, Samuli Laine, and Jaakko Lehtinen.
\newblock Progressive growing of {GAN}s for improved quality, stability, and
  variation.
\newblock In {\em International Conference on Learning Representations}, 2018.

\bibitem{karras2019style}
Tero Karras, Samuli Laine, and Timo Aila.
\newblock A style-based generator architecture for generative adversarial
  networks.
\newblock In {\em Proceedings of the IEEE/CVF Conference on Computer Vision and
  Pattern Recognition}, pages 4401--4410, 2019.

\bibitem{karras2020analyzing}
Tero Karras, Samuli Laine, Miika Aittala, Janne Hellsten, Jaakko Lehtinen, and
  Timo Aila.
\newblock Analyzing and improving the image quality of stylegan.
\newblock In {\em Proceedings of the IEEE/CVF Conference on Computer Vision and
  Pattern Recognition}, pages 8110--8119, 2020.

\bibitem{CelebAMask-HQ}
Cheng-Han Lee, Ziwei Liu, Lingyun Wu, and Ping Luo.
\newblock Maskgan: Towards diverse and interactive facial image manipulation.
\newblock In {\em Proceedings of the IEEE/CVF Conference on Computer Vision and
  Pattern Recognition}, pages 5549--5558, 2020.

\bibitem{leimkuhler2021freestylegan}
Thomas Leimk\"uhler and George Drettakis.
\newblock Freestylegan: Free-view editable portrait rendering with the camera
  manifold.
\newblock {\em ACM Transactions on Graphics (SIGGRAPH Asia)}, 40(6), 2021.

\bibitem{li2020deepfacepencil}
Yuhang Li, Xuejin Chen, Binxin Yang, Zihan Chen, Zhihua Cheng, and Zheng-Jun
  Zha.
\newblock Deepfacepencil: Creating face images from freehand sketches.
\newblock In {\em Proceedings of the 28th ACM International Conference on
  Multimedia}, pages 991--999, 2020.

\bibitem{lin2021barf}
Chen-Hsuan Lin, Wei-Chiu Ma, Antonio Torralba, and Simon Lucey.
\newblock Barf: Bundle-adjusting neural radiance fields.
\newblock In {\em Proceedings of the IEEE/CVF International Conference on
  Computer Vision}, pages 5741--5751, 2021.

\bibitem{liu2021neural}
Lingjie Liu, Marc Habermann, Viktor Rudnev, Kripasindhu Sarkar, Jiatao Gu, and
  Christian Theobalt.
\newblock Neural actor: Neural free-view synthesis of human actors with pose
  control.
\newblock {\em ACM Transactions on Graphics (TOG)}, 40(6):1--16, 2021.

\bibitem{liu2018intriguing}
Rosanne Liu, Joel Lehman, Piero Molino, Felipe Petroski~Such, Eric Frank, Alex
  Sergeev, and Jason Yosinski.
\newblock An intriguing failing of convolutional neural networks and the
  coordconv solution.
\newblock {\em Advances in neural information processing systems}, 31, 2018.

\bibitem{liu2021editing}
Steven Liu, Xiuming Zhang, Zhoutong Zhang, Richard Zhang, Jun-Yan Zhu, and
  Bryan Russell.
\newblock Editing conditional radiance fields.
\newblock In {\em Proceedings of the IEEE/CVF International Conference on
  Computer Vision}, pages 5773--5783, 2021.

\bibitem{liu2015deep}
Ziwei Liu, Ping Luo, Xiaogang Wang, and Xiaoou Tang.
\newblock Deep learning face attributes in the wild.
\newblock In {\em Proceedings of the IEEE international conference on computer
  vision}, pages 3730--3738, 2015.

\bibitem{meng2021gnerf}
Quan Meng, Anpei Chen, Haimin Luo, Minye Wu, Hao Su, Lan Xu, Xuming He, and
  Jingyi Yu.
\newblock Gnerf: Gan-based neural radiance field without posed camera.
\newblock In {\em Proceedings of the IEEE/CVF International Conference on
  Computer Vision}, pages 6351--6361, 2021.

\bibitem{mescheder2018training}
Lars Mescheder, Andreas Geiger, and Sebastian Nowozin.
\newblock Which training methods for gans do actually converge?
\newblock In {\em International conference on machine learning}, pages
  3481--3490. PMLR, 2018.

\bibitem{mescheder2019occupancy}
Lars Mescheder, Michael Oechsle, Michael Niemeyer, Sebastian Nowozin, and
  Andreas Geiger.
\newblock Occupancy networks: Learning 3d reconstruction in function space.
\newblock In {\em Proceedings of the IEEE/CVF Conference on Computer Vision and
  Pattern Recognition}, pages 4460--4470, 2019.

\bibitem{michalkiewicz2019implicit}
Mateusz Michalkiewicz, Jhony~K Pontes, Dominic Jack, Mahsa Baktashmotlagh, and
  Anders Eriksson.
\newblock Implicit surface representations as layers in neural networks.
\newblock In {\em Proceedings of the IEEE/CVF International Conference on
  Computer Vision}, pages 4743--4752, 2019.

\bibitem{mildenhall2020nerf}
Ben Mildenhall, Pratul~P Srinivasan, Matthew Tancik, Jonathan~T Barron, Ravi
  Ramamoorthi, and Ren Ng.
\newblock Nerf: Representing scenes as neural radiance fields for view
  synthesis.
\newblock In {\em European conference on computer vision}, pages 405--421.
  Springer, 2020.

\bibitem{nguyen2019hologan}
Thu Nguyen-Phuoc, Chuan Li, Lucas Theis, Christian Richardt, and Yong-Liang
  Yang.
\newblock Hologan: Unsupervised learning of 3d representations from natural
  images.
\newblock In {\em Proceedings of the IEEE/CVF International Conference on
  Computer Vision}, pages 7588--7597, 2019.

\bibitem{niemeyer2021giraffe}
Michael Niemeyer and Andreas Geiger.
\newblock Giraffe: Representing scenes as compositional generative neural
  feature fields.
\newblock In {\em Proceedings of the IEEE/CVF Conference on Computer Vision and
  Pattern Recognition}, pages 11453--11464, 2021.

\bibitem{orel2021stylesdf}
Roy Or-El, Xuan Luo, Mengyi Shan, Eli Shechtman, Jeong~Joon Park, and Ira
  Kemelmacher-Shlizerman.
\newblock Style{SDF}: {H}igh-{R}esolution {3D}-{C}onsistent {I}mage and
  {G}eometry {G}eneration.
\newblock {\em arXiv preprint arXiv:2112.11427}, 2021.

\bibitem{park2019deepsdf}
Jeong~Joon Park, Peter Florence, Julian Straub, Richard Newcombe, and Steven
  Lovegrove.
\newblock Deepsdf: Learning continuous signed distance functions for shape
  representation.
\newblock In {\em Proceedings of the IEEE/CVF Conference on Computer Vision and
  Pattern Recognition}, pages 165--174, 2019.

\bibitem{park2019semantic}
Taesung Park, Ming-Yu Liu, Ting-Chun Wang, and Jun-Yan Zhu.
\newblock Semantic image synthesis with spatially-adaptive normalization.
\newblock In {\em Proceedings of the IEEE/CVF Conference on Computer Vision and
  Pattern Recognition}, pages 2337--2346, 2019.

\bibitem{peng2021animatable}
Sida Peng, Junting Dong, Qianqian Wang, Shangzhan Zhang, Qing Shuai, Xiaowei
  Zhou, and Hujun Bao.
\newblock Animatable neural radiance fields for modeling dynamic human bodies.
\newblock In {\em Proceedings of the IEEE/CVF International Conference on
  Computer Vision}, pages 14314--14323, 2021.

\bibitem{perez2018film}
Ethan Perez, Florian Strub, Harm De~Vries, Vincent Dumoulin, and Aaron
  Courville.
\newblock Film: Visual reasoning with a general conditioning layer.
\newblock In {\em Proceedings of the AAAI Conference on Artificial
  Intelligence}, volume~32, 2018.

\bibitem{reiser2021kilonerf}
Christian Reiser, Songyou Peng, Yiyi Liao, and Andreas Geiger.
\newblock Kilonerf: Speeding up neural radiance fields with thousands of tiny
  mlps.
\newblock In {\em Proceedings of the IEEE/CVF International Conference on
  Computer Vision}, pages 14335--14345, 2021.

\bibitem{schwarz2020graf}
Katja Schwarz, Yiyi Liao, Michael Niemeyer, and Andreas Geiger.
\newblock Graf: Generative radiance fields for 3d-aware image synthesis.
\newblock {\em Advances in Neural Information Processing Systems},
  33:20154--20166, 2020.

\bibitem{shen2020interfacegan}
Yujun Shen, Ceyuan Yang, Xiaoou Tang, and Bolei Zhou.
\newblock Interfacegan: Interpreting the disentangled face representation
  learned by gans.
\newblock {\em IEEE transactions on pattern analysis and machine intelligence},
  2020.

\bibitem{tewari2020stylerig}
Ayush Tewari, Mohamed Elgharib, Gaurav Bharaj, Florian Bernard, Hans-Peter
  Seidel, Patrick P{\'e}rez, Michael Zollhofer, and Christian Theobalt.
\newblock Stylerig: Rigging stylegan for 3d control over portrait images.
\newblock In {\em Proceedings of the IEEE/CVF Conference on Computer Vision and
  Pattern Recognition}, pages 6142--6151, 2020.

\bibitem{tov2021designing}
Omer Tov, Yuval Alaluf, Yotam Nitzan, Or Patashnik, and Daniel Cohen-Or.
\newblock Designing an encoder for stylegan image manipulation.
\newblock {\em ACM Transactions on Graphics (TOG)}, 40(4):1--14, 2021.

\bibitem{wang2021high}
Tengfei Wang, Yong Zhang, Yanbo Fan, Jue Wang, and Qifeng Chen.
\newblock High-fidelity gan inversion for image attribute editing.
\newblock {\em arXiv preprint arXiv:2109.06590}, 2021.

\bibitem{yu2021plenoctrees}
Alex Yu, Ruilong Li, Matthew Tancik, Hao Li, Ren Ng, and Angjoo Kanazawa.
\newblock Plenoctrees for real-time rendering of neural radiance fields.
\newblock In {\em Proceedings of the IEEE/CVF International Conference on
  Computer Vision}, pages 5752--5761, 2021.

\bibitem{yu2021pixelnerf}
Alex Yu, Vickie Ye, Matthew Tancik, and Angjoo Kanazawa.
\newblock pixelnerf: Neural radiance fields from one or few images.
\newblock In {\em Proceedings of the IEEE/CVF Conference on Computer Vision and
  Pattern Recognition}, pages 4578--4587, 2021.

\bibitem{yu2018bisenet}
Changqian Yu, Jingbo Wang, Chao Peng, Changxin Gao, Gang Yu, and Nong Sang.
\newblock Bisenet: Bilateral segmentation network for real-time semantic
  segmentation.
\newblock In {\em Proceedings of the European conference on computer vision
  (ECCV)}, pages 325--341, 2018.

\bibitem{zhi2021place}
Shuaifeng Zhi, Tristan Laidlow, Stefan Leutenegger, and Andrew~J Davison.
\newblock In-place scene labelling and understanding with implicit scene
  representation.
\newblock In {\em Proceedings of the IEEE/CVF International Conference on
  Computer Vision}, pages 15838--15847, 2021.

\bibitem{zhou2021cips3d}
Peng Zhou, Lingxi Xie, Bingbing Ni, and Qi Tian.
\newblock Cips-3d: A 3d-aware generator of gans based on
  conditionally-independent pixel synthesis.
\newblock {\em arXiv preprint arXiv:2110.09788}, 2021.

\bibitem{zhu2018visual}
Jun-Yan Zhu, Zhoutong Zhang, Chengkai Zhang, Jiajun Wu, Antonio Torralba, Josh
  Tenenbaum, and Bill Freeman.
\newblock Visual object networks: Image generation with disentangled 3d
  representations.
\newblock {\em Advances in neural information processing systems}, 31, 2018.

\bibitem{zhu2020sean}
Peihao Zhu, Rameen Abdal, Yipeng Qin, and Peter Wonka.
\newblock Sean: Image synthesis with semantic region-adaptive normalization.
\newblock In {\em Proceedings of the IEEE/CVF Conference on Computer Vision and
  Pattern Recognition}, pages 5104--5113, 2020.

\end{thebibliography}
}
\end{document}